# Representing Context-Sensitive Knowledge in a Network Formalism: A Preliminary Report

**Tze-Yun Leong**
Clinical Decision Making Group
MIT Laboratory for Computer Science
Cambridge, MA 02139
leong@lcs.mit.edu

## Abstract

Automated decision making is often complicated by the complexity of the knowledge involved. Much of this complexity arises from the context-sensitive variations of the underlying phenomena. We propose a framework for representing descriptive, context-sensitive knowledge. Our approach attempts to integrate categorical and uncertain knowledge in a network formalism. This paper outlines the basic representation constructs, examines their expressiveness and efficiency, and discusses the potential applications of the framework.

## 1 INTRODUCTION

We live in a world which is full of variations and exceptions. Decision making in our daily lives involves skillfully manipulating the myriad of phenomena and carefully analyzing the consequences of each relevant variation or exception. For instance, in the clinical setting, the choice of treatment prescription for a particular disease depends on the general condition of the patient, the presence or absence of other complications, the regimen of other medications being prescribed, *etc.* Hence, to automate the decision making process, there must be a general way to represent the context-sensitive variations of the relevant information.

Research in path-based inheritance in hierarchical systems (Touretzky 1987) and uncertain reasoning with belief networks (Pearl 1988) has shed some light on the characteristics and the complexities of a general framework for reasoning with context-sensitive knowledge. In particular, network or graph representations are found to be very effective in expressing the variations and exceptions involved.

There have been many efforts at integrating categorical or hierarchical knowledge with uncertain knowledge (Lin and Goebel 1990)(Saffiotti 1990)(Yen and Bonissone 1990). No existing framework, however, captures the essence of both, say, the inheritance graph of a specialization or "IS-A" hierarchy, and the conditional dependency graph of a probabilistic network. In other words, current frameworks only allow us to express context-sensitive knowledge either in absolute terms or probabilistically, but not both (Leong 1991b).

In (Leong 1991b), we have identified the different types of information required for supporting dynamic, knowledge-based formulation of decision models in a broad domain. Given a decision problem, dynamic decision modeling involves selecting a subset of concepts and relations from a knowledge base, and assembling them into a closed-world decision model, *e.g.*, an influence diagram (Breese, Goldman and Wellman 1991). Our analysis indicated that an appropriate knowledge base representation would be a network formalism integrating categorical or absolute knowledge and uncertain knowledge in a context-sensitive manner.

We propose such a representation design in this paper. The following information, for example, is expressible in our framework:

### The Royal Elephant Example

*Elephants are gray in color. Royal elephants are a kind of elephants. Royal elephants in Thailand are white in color. Presence of people usually scares away the elephants. But royal elephants are more likely to be found when there are people around. In particular, the King of Thailand always demands the royal elephants in Thailand to follow him everywhere.*

While this piece of (fictitious) information may not seem immediately interesting from the decision making point fo view, it illustrates some important representation requirements that our framework attempts to capture.

First, the different relevant phenomena must be explicitly distinguishable, describable, and capable of supporting reasoning, *e.g.*, elephant, royal elephant, color of elephant, gray, white, Thailand, King of Thailand, *etc.* These descriptions would constitute the basic building blocks of the representation framework.

Second, the different categorical or structural relations among the phenomena must be expressible. Such relations include the specialization or "a kind of" relation, *e.g.*, royal elephant is a kind of elephant, and the decomposition or "part of" relation, the equivalence relation, *etc.*

Similarly, the different uncertain or behavioral relations among the phenomena must be expressible. Instances of such relations, as illustrated in the above example, include those captured in the English phrases: "usually scares away", "more likely to be found", and "always follow."

Lastly, there should be a construct that would capture the context-dependent notions indicated in the Royal Elephant Example: Only the royal elephants in Thailand are white in color, and they can always be found when the King is around. These facts or descriptions are not applicable to royal elephants in general.

Due to its simplicity, we shall refer to the Royal Elephant Example throughout this paper to illustrate the major representation constructs in our framework. Comments on how these constructs are actually being employed will be made whenever appropriate.

In the following sections, we shall describe the components of the proposed framework, and examine some of the motivations behind our design choices. We shall also briefly discuss the typical inferences in automated decision making supported by the framework, and informally assess its potential expressiveness, efficiency, and effectiveness.

## 2 A PARTIAL NETWORK

Figure 1 depicts some relevant parts of the network representation for the Royal Elephant Example in our framework. In the figure, the nodes represent the phenomena or concepts being described, while the links represent the relations among the concepts. Only one type of categorical or structural relations is displayed: specialization (AKO). Three types of uncertain or behavioral relations are displayed: cause (c), positive-influence (+), and negative-influence (-). A third type of relations, the context (CXT) relation, induces a hypergraph on the network; the transitive-closure of the context relation of a concept constitutes its description. The (#) and (#*) signs in the figure should be read as: "of", e.g., "King of Thailand", "Presence of King of Thailand"; the (#*) sign is simply an abbreviation of an implicit chain of the (#) signs.

In contrast to early semantic networks with ad-hoc relations, to term-subsumption languages with only subsumption (IS-A) relations, and to belief networks with only probabilistic relations, our representation design accommodates a spectrum of different relations with well-defined, though not necessarily formal semantics. We shall now look at the different components in more details.

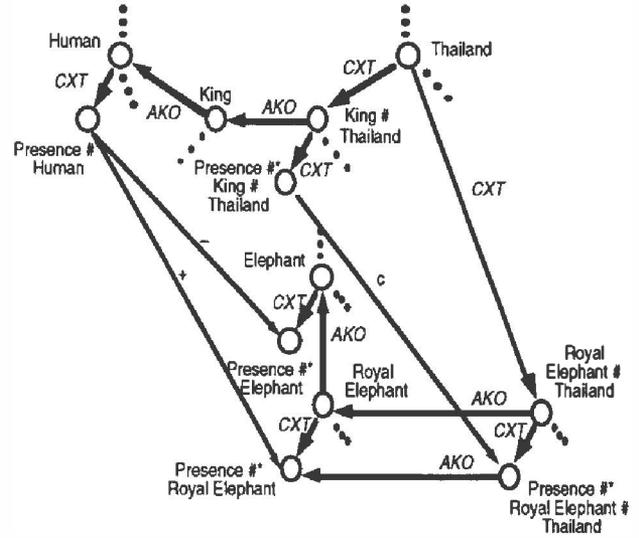

Figure 1: Partial Network Representation of the Royal Elephant Example

## 3 REPRESENTATION OF CONCEPTS

In our framework, a *concept* is an intensional description of the relational interpretation of an object, a state, a process, or an attribute of these phenomena. In other words, a concept reflects the salient features of the underlying phenomenon through a set of *interactions*, i.e., correlational, influential, or causal relations with other concepts. For example, the concept royal elephant might comprise the following relations[1]:

- "age of royal elephant positively-influences length of teeth of royal elephant",
- "gender of royal elephant associates-with size of royal elephant", *etc*.

In these relations, concepts such as age of royal elephant, teeth of royal elephant, gender of royal elephant, and size of royal elephants are related to royal elephant via the *context* or CXT relation; they are called the *properties* of royal elephant, and in turn may have their own properties, *e.g.*, length of teeth of royal elephant is a property of teeth of royal elephant.

The description of a concept, *i.e.*, its properties and the interactions among them, may be constrained by a set of *categorizers*. A categorizer is a categorical or class relationship which establishes a unique perspective for describing one concept in terms of another. For example, asserting the relation: "royal elephant is a kind of elephant" in the description of royal elephant implies that its properties and their corresponding interactions may have been

---

[1] Relations such as: "color of royal elephant is white" are special case to this characterization, and can be handled in a similar way.

*inherited*, in a particular manner, from those of elephant. A partial network representation of the concept royal elephant is shown in Figure 2.

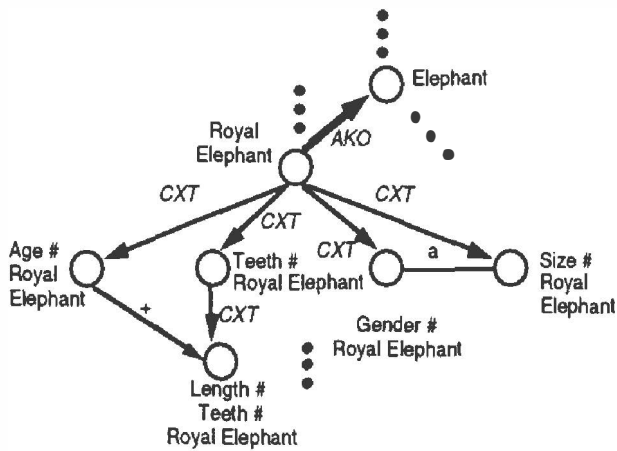

Figure 2: Partial Description of the Royal Elephant Concept

The rationale behind our design is discussed in detail in (Leong 1991a). In essence, the different relations defined reflect the characteristics of the knowledge involved in supporting dynamic decision modeling.

The interactions capture behavioral relations with varying degrees of certainty among the concepts; these relations support the task of identifying information with varying degrees of significance in a particular situation. For instance, in deciding a treatment plan for a disease, the decision maker might wish to consider other events or conditions that affect or are affected by the disease, *e.g.,* its potential causes, its symptoms, its complications, *etc.* The relevance of these related events is discriminated according to the certainty or "strength" of their interactions with the disease.

The categorizers capture structural relations among the concepts; these relations support the task of reasoning at multiple levels of details in decision modeling. For instance, given the presence of a disease, say pneumonia, a decision maker might wish to prescribe treatment after deciding which particular subtype of pneumonia is actually present. The possible subtypes of pneumonia can be found by tracing the concepts related to pneumonia via the specialization (AKO) relation.

One important component of our representation design is the context (CXT) relation. This unique relation is neither behavioral nor structural, instead, it can be regarded as a higher-order relation that constrains the interpretations of all other relation types in the framework. Explicit encoding of the CXT relations provides a general mechanism to describe the concepts, in terms of their other types of relations among each other, in a context-sensitive manner. Such information is crucial for supporting decision modeling in "abnormal" or "non-general" situations. For instance, in the decision problem above, if a second disease, say Acquired Immune-deficiency Syndrome (AIDS) is present, the decision maker should consider some subtypes of pneumonia which are different from those being considered in the absence of AIDS.

### 3.1 THE CONTEXT RELATION

Intuitively, the *context* or CXT relation delimits the "scope" of the description of a concept in a network. All concepts in our framework are denoted in terms of the CXT relation.

In general, all concepts reachable from a particular concept, say C, via the CXT relations in the network are in the description of C. A concept directly related to C via the CXT relation is a *property* of C, denoted as (P # C), *e.g.,* (King # Thailand), read: King of Thailand, and (Royal Elephant # Thailand), read: Royal Elephant of Thailand. The properties of a concept include its inherent qualities, characteristics, and other relevant concepts that constitute its description.

It follows that every concept is defined in some context. In other words, all concepts can be expressed in the form of (a # b). In this tuple notation, a is the "basic identity" of the concept, and b is the "context" in which the concept is defined; both entries are concepts themselves. The basic identify of a concept is the most accurate general description of the concept. The context specifies the condition in which the description of the denoted concept is valid, and allows this description to vary, if necessary, from the basic identity. There is a special concept, denoted as T, which is defined to be itself; any concept defined in the context of T is in the universal context, i.e., valid in general. For example, the concepts human, elephant, royal elephant, etc., are actually denoted as (Human # T), (Elephant # T), (Royal Elephant # T), and so forth. For simplicity, we shall omit the universal context in our notations in this paper.

The tuple notation allows concepts to be "chained" to form a new concept, analogous to the "role chaining" notion in KL-ONE (Brachman and Schmolze 1985). For instance, ((Color # Royal Elephant) # Thailand) is a concept. The chaining expression is associative, and the embedded parentheses are usually omitted.

The CXT relation, therefore, induces a "context tree" among all the concepts defined in the knowledge base, with the universal concept T as the root. This context hierarchy serves two purposes: First, as we shall see below, it allows expression of context-sensitive description of a concept in terms of its categorical and uncertain relations with other concepts. Second, it serves as a focusing mechanism because, as we have noted earlier, every subtree in the hierarchy contains all the relevant concepts in the description of the particular concept at the root of the subtree.

### 3•2 BEHAVIORAL RELATIONS: INTERACTIONS

An interaction is a "behavioral" relationship between two or more concepts. In the decision modeling context, the interactions can be described in terms of English words such as "causes," "alleviates," "indicates," *etc.*, in one extreme; they can also be expressed as numeric conditional probabilities between two or more concepts in another extreme. To balance between intuitive expressiveness and semantic precision, our definitions integrate a temporal ordering notion and a qualitative probabilistic interpretation.

Each interaction in our framework has two components: *temporal precedence*, with "known" or "unknown" as values, and *qualitative probabilistic influence* (Wellman 1990b), with "positive", "negative", or "unknown" as values. Different additive combinations of these values allow us to express the behavioral relationships across a spectrum of uncertainty.

The interpretations for the temporal precedence values are straightforward. The qualitative probabilistic influence values, in a nutshell, are defined as follows: if a concept $C_1$ positively/negatively influence another concept $C_2$, then 1) for binary concepts $C_1$ and $C_2$, the presence of $C_1$ increases/decreases the probability of the presence of $C_2$, with all other things being unchanged; and 2) for continuous concepts $C_1$ and $C_2$, higher values of $C_1$ increase/decrease the probability of higher values of $C_2$, again with all other things being unchanged. The detailed definitions can be found in (Leong 1991a) and (Wellman 1990b). Table 1 depicts the four types of interactions defined between any two concepts $c_1$ and $c_2$: association, precedence, influence, and cause/inhibition.

Table 1: Types of Interactions

| Interaction Type | Network Representation | Temporal Precedence | Qualitative Probabilistic Influence |
|---|---|---|---|
| Association | $(C_1) \xrightarrow{a} (C_2)$ | Unknown | Unknown |
| Precedence | $(C_1) \xrightarrow{p} (C_2)$ | Known | Unknown |
| Influence | $(C_1) \xrightarrow{+/-} (C_2)$ | Unknown | Positive/Negative |
| Cause/Inhibition | $(C_1) \xrightarrow{c/i} (C_2)$ | Known | Positive/Negative |

Interactions with known temporal ordering can only be used to describe concepts that represent "events" in the world. The temporal ordering also constrains the possible interactions among a set of concepts. For example, if concept $A$ precedes concept $B$, either by direct or indirect interaction, it is not allowed to assert an influence from $B$ to $A$.

In the above definitions, interpreting causation/inhibition as positive/negative probabilistic influences with known temporal precedence is consistent with the standard definition of probabilistic causality with temporal ordering, as proposed by Suppes (Suppes 1970).

With reference to Figure 1, in the Royal Elephant Example, the statements: "presence of people usually scares away the elephants", "royal elephants are more likely to be found when there are people around", and "the King of Thailand always demands the royal elephants in Thailand to follow him everywhere" can be expressed as the following interactions:

- <u>Presence of human</u> negatively-influences <u>presence of elephant</u>
- <u>Presence of human</u> positively-influences <u>presence of royal elephant</u>
- <u>Presence of King of Thailand</u> causes <u>presence of royal elephant of Thailand</u>

### 3•3 STRUCTURAL RELATIONS: CATEGORIZATIONS

A categorizer is a binary relation that groups concepts, according to their descriptions, into a *categorization*. By knowing the position of a particular concept with respect to another concept in a categorization, we can infer the description, *i.e.*, the properties and their corresponding interactions of the former from the latter in a particular manner. Examples of categorizers, as defined in our framework, include the specialization or "a kind of" (AKO) relation, the decomposition or "part of" (PARTOF) relation, the equivalence (EQV) relation, and the structural-copy (SC) relation. In the Royal Elephant Example, the relevant categorical relationships are:

- <u>Royal elephant</u> is a kind of <u>elephant</u>
- <u>Royal elephant of Thailand</u> is a kind of <u>royal elephant</u>
- <u>King</u> is a kind of <u>human</u>
- <u>King of Thailand</u> is a kind of <u>king</u>

A concept can be involved in multiple categorizations, *e.g.*, <u>teeth of royal elephant</u> is part of <u>royal elephant</u>, and also is a kind of <u>organ of animal</u>. A set of conventions, based on subgraphs copying and references updating, are defined for each categorizer for proper inheritance of concept descriptions. It is currently assumed that the descriptions inherited in different categorizations of the concept are consistent[2].

The specialization, decomposition, and equivalence rela-

---

[2] As we shall discuss in the potential application of the framework, this assumption is quite reasonable.

tions defined in our framework are in accordance with the conventional or common definitions in the knowledge representation literature. We shall not repeat the definitions here. The context sensitive nature of our framework, however, calls for the formalization of a new categorical relation, structural-copy (SC). The SC relation can be viewed as a unidirectional "reference" relation. This relation is not explicitly demonstrated in the Royal Elephant Example, but we could easily extend the scenario as follows:

**The Royal Elephant Example (cont.)**

*As mentioned earlier, the royal elephants in Thailand are present whenever the King of Thailand makes a public appearance. Those royal elephants with pink tails in Thailand are always selected as the King's rides.*

In this case, we shall define a concept pink-tail royal elephant of Thailand, which is kind of royal elephant of Thailand. There is also a need to define another concept called ride of King of Thailand, with which we associate the description for a typical ride for a king, *e.g.,* the type of saddle mounted, decorations, *etc.* But the ride of King of Thailand is also a pink-tail royal elephant of Thailand, in the sense that the description of the latter can be directly used to describe the former. Note that this is not a specialization relation, *i.e.,* the ride of King of Thailand is not a kind of pink- tail royal elephant of Thailand; the two concepts are actually descriptions of the same object under different circumstances[3] Therefore, it is much more natural to define ride of King of Thailand as a structural-copy of pink-tail royal elephant of Thailand.

In general, if a concept $A$ is a structural copy of another concept $B$, then the description of $B$ is *visible* in the description of $A$. In other words, the properties and their corresponding interactions of $B$ may be directly used in the description of $A$, with the appropriate updated references. For instance, the property teeth of ride of King of Thailand is directly copied from the corresponding property teeth of pink-tail royal elephant of Thailand. In the planned implementation of the framework, we do not have to specify this description in the definition of ride of King of Thailand; as long as the SC relation between the two concepts is asserted, the corresponding structure should be automatically copied when the knowledge base is constructed.

The SC relation is irreflexive, antisymmetric, and transitive. Intuitively, the SC relation provides a means for different concepts to share description under different constraints or situations, *e.g.,* in Thailand. These extra constraints or situations are captured in the CXT relations of the concepts involved.

## 4  STRUCTURE OF KNOWLEDGE BASE

So far we have outline the basic representation constructs in our framework. By adopting a descriptive approach to concept definition, we have developed a set of categorical relations and a set of uncertain relations among the concepts. These relations are further constrained by the CXT relation to capture context-sensitive information in a uniform way.

From the network perspective, each type of relations defined in our framework imposes a set of multiply connected directed graphs on the concepts. In particular, the CXT relation hierarchy forms a single directed tree on all the concepts in the knowledge base. This imposed regularity on the knowledge base, we believe, would facilitate the efficiency of the inferences supported.

A major assumption that allows us to take advantage of the network interpretation of the framework for supporting inferences is that all the relation links in concept descriptions, including those that are inherited, are fully established when the concepts are defined. In other words, all the concept descriptions are "pre-compiled", and no "run-time" definition is allowed. This strong assumption has simplified the representation design process, but will likely to be eliminated as we progress to explore more complicated issues and improve our design in the future.

As mentioned earlier, a subtree in the context hierarchy is built for each concept defined., with its properties in turn as the branches or subtrees of this subtree. Given that all the relation links are fully established for each concept, at first glance, the possible "chaining" of the CXT relation would lead to an exponential explosion in he number of definable or derivable concepts.

Indeed, the number of distinct concepts that can be formed from an initial set of $n$ context-free concepts, *i.e.,* concepts defined in the universal concept T, are of $O(n!)$ or $O(n^n)$. The actual bound for the knowledge is actually infinite if we allow a concept to appear more than once in a CXT chain, *e.g.,* (child # child # child #.... # King # Thailand). The space needed for the knowledge base could possibly be huge. We believe, however, the situation is not that serious because, in general, many of the CXT chaining combinations do not make sense; the CXT hierarchy is usually sparse.

## 5  INFERENCES SUPPORTED

Two classes of powerful inferences, *inheritance* and *recognition,* are usually supported in hierarchical knowledge representation systems of the *semantic networks* family.

---

[3.] Another more realistic example is the concept: complication of AIDS, which is usually another disease, say Pneumocystis carinii pneumonia (PCP), or a physiological state. In this case, the description of PCP can be used to describe the concept, in addition to its description of being a complication.

The presence of conflicting concept descriptions gives rise to the exceptions and multiple inheritance problems in inheritance, and the partial matching problem in recognition (Shastri 1989). Since we assume our knowledge base is a fully established network of concept descriptions, we do not anticipate most of the difficulties that research in inheritance theory or default reasoning (Touretzky, Horty and Thomason 1987) encounters. As compared to these systems for supporting commonsense reasoning, however, only a restricted set of inferences are provided in our framework.

All the knowledge in our knowledge base is currently assumed to be pre-compiled; any conflicts or inconsistencies would have been resolved, either by the conventions specified in the relational semantics or by consulting the user, when the network is constructed. The multiple inheritance problem in our framework is therefore addressed when the knowledge base is constructed; the exceptions are handled by explicitly specifying the CXT relations in a uniform way. There is no run-time support for inheritance inferences.

On the other hand, our framework is equipped to handle a restricted class of the recognition problem; these problems can be reduced to the simpler problem of finding a path in a particular network imposed by a relation type, and then interpreting the indirect relation between the concepts at the beginning and the end of the path.

### 5•1 INDIRECT INTERACTIONS

There are two forms of indirections for interactions: *interaction chains* and *parallel interactions*. An example of the former scenario is as follows:

- Presence of human negatively-influences presence of elephant
- Presence of elephant positively-influences presence of mouse

A relevant query would be: What is the interaction between presence of human and presence of mouse†?

Similarly, an example of the latter scenario is as follows:

- Presence of King of Thailand causes presence of royal elephant of Thailand
- Presence of King of Thailand positively-influences presence of cat of Thailand
- Presence of royal elephant of Thailand positively-influences presence of mouse of Thailand
- Presence of cat of Thailand negatively-influences presence of mouse of Thailand

A relevant query would be: What is the net interaction between King of Thailand and mouse of Thailand?

Table 2 defines the indirect effects of the interactions. The " $\otimes$ " operator is for combining interaction chains and the " $\oplus$ " operator is for combining parallel interactions. The definitions are consistent with the operators for combining *influence chains* and *parallel influences* in qualitative probabilistic networks (QPNs). The corresponding operations are commutative, associative, and distributive, just like ordinary multiplication and addition (Wellman 1990b). The tables are indexed from interaction entries in "row" then "column", and the net interaction is read from their intersection. For example, $- \otimes i = +; + \oplus c = c$.

Table 2: Indirect Effects of Interactions

| $\otimes$ | a | p | + | - | c | i | $\oplus$ | a | p | + | - | c | i |
|---|---|---|---|---|---|---|---|---|---|---|---|---|---|
| a | a | a | a | a | a | a | a | a | p | a | a | p | p |
| p | a | p | a | a | p | p | p | p | p | p | p | p | p |
| + | a | a | + | - | + | - | + | a | p | + | a | c | p |
| - | a | a | - | + | - | + | - | a | p | a | - | p | i |
| c | a | p | + | - | c | i | c | p | p | c | p | c | p |
| i | a | p | - | + | i | c | i | p | p | p | i | p | i |

### 5•2 INDIRECT CATEGORIZATIONS

Determining the relationship between two concepts in a particular categorization is straightforward, involving simply checking whether one concept is in the transitive closure of the other. The context-sensitive nature of our framework further allows, for example, the following types of inferences to be drawn on the categorizations:

- Elephant is a kind of animal
- Teeth is a kind of organ

We can conclude that:

- Teeth of elephant is a kind of organ of animal

The detailed definition of such inferences is again documented in (Leong 1991a). In the specialization hierarchy, this definition is analogous to the idea of *derivative subclassification* in OWL (Hawkinson 1975). The inferences supported in our framework are generalized to all other categorical relations defined as well.

## 6 SUPPORTING DECISION MAKING

We have seen how the relevant information in the Royal Elephant Example can be adequately captured in our representation framework. We shall now examine how the represented knowledge can be used to support dynamic formulation of a decision model.

The decision-analytic approach to decision making can be viewed as a five-step process: 1) Background information characterization; 2) domain context establishment; 3) decision problem formulation; 4) decision model construction; and 5) decision model evaluation.

To support the above decision making process, the following general types of queries are involved (Leong 1991b), with the parameters in the angular brackets denoting the relations defined in our framework:

- (Q1) Does concept $\underline{A}$ relate to concept $\underline{B}$ by <categorizer>?
- (Q2) What are the concepts related to concept $\underline{A}$ by <categorizer>?
- (Q3) Does concept $\underline{A}$ relate to concept $\underline{B}$ by <interaction>?
- (Q4) What are the concepts related to concept $\underline{A}$ by <interaction>?

For example, consider the following scenario:

### The Tourist's Decision Problem

*A tourist in Thailand had a very expensive camera. One day very early in the morning, He heard from the radio that an elephant was spotted in a nearby shopping mall. He would really like to take a picture of a Royal Elephant, but the radio report did not mention what type of elephant it was. The decision is whether or not the tourist should bring his camera to the shopping mall, given a substantial chance that the camera could be stolen.*

A target decision model for the Tourist's Decision Problem is shown in Figure 3.

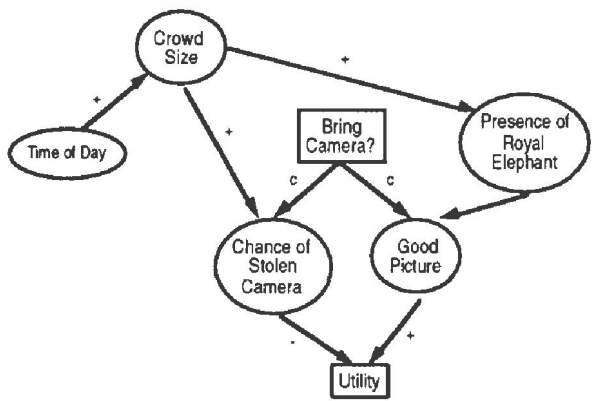

Figure 3: A QPN for the Tourist's Decision Problem

Given the problem specification, and a knowledge base containing all the relevant information about elephants and a tourist's life in Thailand, the decision maker could formulate the above decision model by posing a series of queries to the knowledge base. Some examples of these queries, are as follows:

- What are the concepts related to Elephant by specialization?
- What are the concepts that positively-influence chance of stolen camera?
- *etc.*

To evaluate the decision model, the decision maker would in turn pose a series of queries to the constructed decision model as follows:

- Does bring-camera? relate to utility by positive-influence?
- Does bring-camera? relate to utility by negative-influence?
- *etc.*

All these queries are of the general forms Q1 to Q4 as defined above. As illustrated in the previous section, these queries are supported by the inferences (direct or indirect interactions and categorizations) provided in our framework.

The built-in context-sensitive nature of the representation provides the decision maker with a general way of accessing variations in the domain information. For example, if the Tourist's Decision Problem is posed in a country other than Thailand, the resulting target decision model might be different because the royal elephants there, if present, might be scared of people. The same set of queries, however, would be used by the decision maker to construct this new decision model

## 7  DISCUSSION AND CONCLUSION

In this paper, we have briefly discussed the motivation and the design approach for a representation framework that integrates categorical knowledge and uncertain knowledge in a context-sensitive manner. Our design is based on a network formalism which facilitates the interpretation and the manipulation of the inheritance problem in the various relations being modelled. By examining how the information in the Royal Elephant Example can be represented, we have demonstrated the expressiveness of our framework. We have also argued that this expressiveness is adequate for capturing many interesting phenomena essential for supporting automated decision making.

Efficiency, *i.e.*, how easily can the knowledge be accessed in the framework, is demonstrated through a set of indirect inference definitions. With these inferences, a restricted class of the recognition problem can be reduced to a path-finding problem. We postulate that instead of the NP-complete *classification* mechanism being supported in most existing term-subsumption languages or representation systems, simple path-finding graph algorithms of polynomial time complexity are adequate for our purpose. More rigorous analysis, however, needs to be done to substantiate this claim.

We would like to conclude the informal evaluation of our framework by examining its *effectiveness*, *i.e.*, how well it supports the applications it is designed for. In this case, the intended application is for supporting dynamic formulation of decision models in automated decision analysis. We have briefly sketched how the framework supports the process with the Tourist's Decision Problem example. In practice, we have also briefly examined this issue by hand-building and hand-evaluating a small test knowledge base in the domain of opportunistic pulmonary infections with suspected AIDS (Leong 1991a); the results are promising. Unfortunately, a rigorous evaluation is impossible until we have an implemented system, which is planned for the near future.

We believe our representation framework is applicable in some other problem solving tasks as well. The restricted set of inferences provided, however, renders it unsuitable for supporting more general recognition problems. Moreover, we have only dealt with concept types and relation types in our framework; concept instances and relation instances are not currently handled. Therefore, any inferences involving instances are not currently addressed, *e.g.*, we would not know what to do with a concept Clyde, which is an instance of royal elephant.

Given the pre-compiled nature of the knowledge base, one might also wonder how easily new information or changes can be incorporated into the intricate network structure. This problem might be alleviated by the appropriate use of delayed evaluation or selective expansion techniques, but we have yet to consider the options carefully to substantiate the claim. This would be a major component to be worked out and considered in evaluating the effectiveness of the implemented framework in future.

In conclusion, while there is definitely much more to be accomplished in this project, we believe we have established the essential components of the proposed representation framework. We have also demonstrated its potentials in facilitating automated decision making under uncertainty. Future agenda for this work include: 1) Implementation of the representation system; 2) formal evaluation of the framework in actual use; 3) refining the relational definitions in the framework; 4) extending the framework to handle concept and relation instances, and 5) development of a set of techniques for efficient incorporation of changes into the knowledge base.

### Acknowledgments

The author wishes to thank Peter Szolovit and Mike Wellman for many helpful discussions and comments on the content of this paper, and the anonymous referees for suggestions on the presentation of this paper. This research was supported by the National Institutes of Health grant no. 5 R01 LM04493 from the National Library of Medicine.